\documentclass{article}

\PassOptionsToPackage{numbers, compress}{natbib}

\usepackage[final]{neurips_2024}

\usepackage[utf8]{inputenc} 
\usepackage[T1]{fontenc}    
\usepackage{hyperref}       
\usepackage{url}            
\usepackage{booktabs}       
\usepackage{amsfonts}       
\usepackage{nicefrac}       
\usepackage{microtype}      
\usepackage{xcolor}         
\usepackage{graphicx}
\usepackage{wrapfig}
\usepackage{float}
\usepackage{CJKutf8}        

\title{Emergence of Implicit World Models\\ from Mortal Agents}
\author{%
  Kazuya Horibe$^*$ \\
  RIKEN \\
  \texttt{horibe.289@gmail.com}\\
  \And
  Naoto Yoshida$^*$ \\ 
  Kyoto University\\
  \texttt{yoshida.naoto.8x@kyoto-u.ac.jp} \\
  \thanks{Both authors contributed equally to this manuscript.}
}

\begin{document}
\begin{CJK}{UTF8}{min}

\maketitle

Life possesses agency and behaves autonomously \cite{moreno2015biological,di2014enactive, tomasello2022evolution}.
Agency refers to the ability to autonomously set goals based on intrinsic motivation (IM) and act toward achieving them. 
Life, by autonomously setting its own goals, is able to proactively respond to unknown situations and unpredictable events, and adjust its behavior using feedback from the environment. 
When attempting to mimic the agency of life, it is crucial for artificial agents to intrinsically set their own goals. 
Intrinsic goal setting has been explored through concepts like prediction information maximization \cite{ay2008predictive,hohwy2013predictive,parr2022active}, empowerment maximization \cite{klyubin2005empowerment,salge2014empowerment}, curiosity-driven learning \cite{schmidhuber1991possibility,oudeyer2007intrinsic,schmidhuber2010formal,burda2018large}, and novelty-based learning \cite{mouret2011novelty,mouret2015illuminating,burda2019exploration}. 
In many of these IM approaches, however, researchers explicitly design motivations, such as ``novelty is good'' in novelty-based learning. 
As a result, the complete internalization of goals within artificial agents has not yet been fully achieved, and flexible adaptation to the environment based on autonomous goal setting remains a challenge.

The most fundamental goal of life is to avoid death. 
Avoiding death means maintaining a state of being alive, that is, possessing homeostasis \cite{cannon1939wisdom,ashby1952design,billman2020homeostasis}, which involves acquiring energy from external sources and keeping one's internal state within a certain range. 
The homeostasis is based on the objective of sustaining the very existence (being) of the self. 
The characterization of life based on the goal of maintaining the persistence of being was proposed as autopoiesis by Maturana and Varela \cite{maturana1980autopoiesis} and later extended by Barandiaran et al. to define agency \cite{barandiaran2009defining}. 
Autopoiesis is a process by which life, driven by the meta-goal of preserving its own existence (being), autonomously sets multiple internalized motivations, such as acquiring energy or escaping predators, and generates open-ended behaviors to achieve them (Appendix \ref{sec:review}) \cite{di2003organismically}.
This suggests the renaissance of research stance that the existence of the agent itself and the extrinsic motivation (EM) to maintain it precedes the agent’s IM, which this stance akin to the perspective of classical suggestions, such as Parisi's internal robotics \cite{parisi2004internal}, and Di Paolo's approach to the homeostatic adaptation using evolutionary optimizations \cite{di2003organismically}. 

A theoretical framework where homeostasis as the core of the EM is known as homeostatic reinforcement learning (homeostatic RL) in computational neuroscience \cite{keramati2011reinforcement,keramati2014homeostatic,hulme2019neurocomputational}. 
By combining deep RL \cite{mnih2015human,schulman2017proximal}, recent studies have reported the emergence of various goal-directed behaviors \cite{yoshida2024emergence,yoshida2024synthesising}. 
These results suggest a possibility of the emergence of highly adaptive process of the artificial systems from our perspective, such as world models and IM (Appendix \ref{sec:gabst}) \cite{ha2018worldmodels,hafner2020mastering,hafner2023mastering,oudeyer2007typology,haber2018learning,sekar2020planning,kauvar2023curious}. In this paper, by combining meta-RL \cite{wang2016learning,wang2018prefrontal} and deep homeostatic RL, we hypothesize the possibility of explaining such IMs as an emergent property of homeostatic systems, together with world models. The further discussion in Appendix \ref{sec:meta_hrl} suggests that including recurrent neural networks (RNN) in homeostatic RL may naturally lead to the meta-learning ability of agents. Furthermore, as reported by Wang et al. \cite{wang2016learning,wang2018prefrontal}, computational experiments have shown that, even though all of the agent architecture and optimization are carried out in {\it model-free}, such meta-RL agents behave like {\it model-based} \cite{wang2018prefrontal}. This suggests that the agent acquires a process for implicitly constructing a model of the environment ({\it implicit} world model) within the unstructured network, and uses it for learning and exploration.



By conducting meta-RL based on the unified EM (homeostasis), we propose a ``mortal agent'' that can open-endedly generate IMs and world models according to the agent-environment coupling.

\newpage

\bibliographystyle{unsrt} 
\bibliography{main}

\begin{thebibliography}{10}

\bibitem{moreno2015biological}
Alvaro Moreno and Matteo Mossio.
\newblock {\em Biological autonomy: A Philosophical and Theoretical Enquiry}.
\newblock Springer, 2015.

\bibitem{di2014enactive}
Ezequiel Di~Paolo and Evan Thompson.
\newblock The enactive approach.
\newblock In {\em The Routledge handbook of embodied cognition}, pages 68--78. Routledge, 2014.

\bibitem{tomasello2022evolution}
Michael Tomasello.
\newblock {\em The evolution of agency: Behavioral organization from lizards to humans}.
\newblock MIT Press, 2022.

\bibitem{ay2008predictive}
Nihat Ay, Nils Bertschinger, Ralf Der, Frank G{\"u}ttler, and Eckehard Olbrich.
\newblock Predictive information and explorative behavior of autonomous robots.
\newblock {\em The European Physical Journal B}, 63:329--339, 2008.

\bibitem{hohwy2013predictive}
Jakob Hohwy.
\newblock {\em The predictive mind}.
\newblock OUP Oxford, 2013.

\bibitem{parr2022active}
Thomas Parr, Giovanni Pezzulo, and Karl~J Friston.
\newblock {\em Active inference: the free energy principle in mind, brain, and behavior}.
\newblock MIT Press, 2022.

\bibitem{klyubin2005empowerment}
Alexander~S Klyubin, Daniel Polani, and Chrystopher~L Nehaniv.
\newblock Empowerment: A universal agent-centric measure of control.
\newblock In {\em 2005 ieee congress on evolutionary computation}, volume~1, pages 128--135. IEEE, 2005.

\bibitem{salge2014empowerment}
Christoph Salge, Cornelius Glackin, and Daniel Polani.
\newblock Empowerment--an introduction.
\newblock {\em Guided Self-Organization: Inception}, pages 67--114, 2014.

\bibitem{schmidhuber1991possibility}
J{\"u}rgen Schmidhuber.
\newblock A possibility for implementing curiosity and boredom in model-building neural controllers.
\newblock In {\em Proc. of the international conference on simulation of adaptive behavior: From animals to animats}, pages 222--227, 1991.

\bibitem{oudeyer2007intrinsic}
Pierre-Yves Oudeyer, Frdric Kaplan, and Verena~V Hafner.
\newblock Intrinsic motivation systems for autonomous mental development.
\newblock {\em IEEE transactions on evolutionary computation}, 11(2):265--286, 2007.

\bibitem{schmidhuber2010formal}
J{\"u}rgen Schmidhuber.
\newblock Formal theory of creativity, fun, and intrinsic motivation (1990--2010).
\newblock {\em IEEE transactions on autonomous mental development}, 2(3):230--247, 2010.

\bibitem{burda2018large}
Yuri Burda, Harri Edwards, Deepak Pathak, Amos Storkey, Trevor Darrell, and Alexei~A Efros.
\newblock Large-scale study of curiosity-driven learning.
\newblock {\em arXiv preprint arXiv:1808.04355}, 2018.

\bibitem{mouret2011novelty}
Jean-Baptiste Mouret.
\newblock Novelty-based multiobjectivization.
\newblock In {\em New Horizons in Evolutionary Robotics: Extended Contributions from the 2009 EvoDeRob Workshop}, pages 139--154. Springer, 2011.

\bibitem{mouret2015illuminating}
Jean-Baptiste Mouret and Jeff Clune.
\newblock Illuminating search spaces by mapping elites.
\newblock {\em arXiv preprint arXiv:1504.04909}, 2015.

\bibitem{burda2019exploration}
Yuri Burda, Harrison Edwards, Amos Storkey, and Oleg Klimov.
\newblock Exploration by random network distillation.
\newblock In {\em Seventh International Conference on Learning Representations}, pages 1--17, 2019.

\bibitem{cannon1939wisdom}
Walter~Bradford Cannon.
\newblock {\em The wisdom of the body}.
\newblock Norton \& Co., 1939.

\bibitem{ashby1952design}
W~Ross Ashby.
\newblock {\em Design for a brain.}
\newblock Wiley, 1952.

\bibitem{billman2020homeostasis}
George~E Billman.
\newblock Homeostasis: the underappreciated and far too often ignored central organizing principle of physiology.
\newblock {\em Frontiers in physiology}, 11:200, 2020.

\bibitem{maturana1980autopoiesis}
Humberto~R Maturana and Francisco~J Varela.
\newblock {\em Autopoiesis and cognition: The realization of the living}.
\newblock Springer, 1980.

\bibitem{barandiaran2009defining}
Xabier~E Barandiaran, Ezequiel Di~Paolo, and Marieke Rohde.
\newblock Defining agency: Individuality, normativity, asymmetry, and spatio-temporality in action.
\newblock {\em Adaptive behavior}, 17(5):367--386, 2009.

\bibitem{di2003organismically}
Ezequiel~A Di~Paolo.
\newblock Organismically-inspired robotics: homeostatic adaptation and teleology beyond the closed sensorimotor loop.
\newblock {\em Dynamical systems approach to embodiment and sociality}, pages 19--42, 2003.

\bibitem{parisi2004internal}
Domenico Parisi.
\newblock Internal robotics.
\newblock {\em Connection science}, 16(4):325--338, 2004.

\bibitem{keramati2011reinforcement}
Mehdi Keramati and Boris~S Gutkin.
\newblock A reinforcement learning theory for homeostatic regulation.
\newblock In {\em Advances in Neural Information Processing Systems}, pages 82--90, 2011.

\bibitem{keramati2014homeostatic}
Mehdi Keramati and Boris Gutkin.
\newblock Homeostatic reinforcement learning for integrating reward collection and physiological stability.
\newblock {\em Elife}, 3:e04811, 2014.

\bibitem{hulme2019neurocomputational}
Oliver~J Hulme, Tobias Morville, and Boris Gutkin.
\newblock Neurocomputational theories of homeostatic control.
\newblock {\em Physics of life reviews}, 31:214--232, 2019.

\bibitem{mnih2015human}
Volodymyr Mnih, Koray Kavukcuoglu, David Silver, Andrei~A Rusu, Joel Veness, Marc~G Bellemare, Alex Graves, Martin Riedmiller, Andreas~K Fidjeland, Georg Ostrovski, et~al.
\newblock Human-level control through deep reinforcement learning.
\newblock {\em Nature}, 518(7540):529--533, 2015.

\bibitem{schulman2017proximal}
John Schulman, Filip Wolski, Prafulla Dhariwal, Alec Radford, and Oleg Klimov.
\newblock Proximal policy optimization algorithms.
\newblock {\em arXiv preprint arXiv:1707.06347}, 2017.

\bibitem{yoshida2024emergence}
Naoto Yoshida, Tatsuya Daikoku, Yukie Nagai, and Yasuo Kuniyoshi.
\newblock Emergence of integrated behaviors through direct optimization for homeostasis.
\newblock {\em Neural Networks}, 177:106379, 2024.

\bibitem{yoshida2024synthesising}
Naoto Yoshida, Hoshinori Kanazawa, and Yasuo Kuniyoshi.
\newblock Synthesising integrated robot behaviour through reinforcement learning for homeostasis.
\newblock {\em bioRxiv}, pages 2024--06, 2024.

\bibitem{ha2018worldmodels}
David Ha and J{\"u}rgen Schmidhuber.
\newblock Recurrent world models facilitate policy evolution.
\newblock In {\em Advances in Neural Information Processing Systems}, pages 2451--2463, 2018.

\bibitem{hafner2020mastering}
Danijar Hafner, Timothy~P Lillicrap, Mohammad Norouzi, and Jimmy Ba.
\newblock Mastering atari with discrete world models.
\newblock In {\em International Conference on Learning Representations}, 2020.

\bibitem{hafner2023mastering}
Danijar Hafner, Jurgis Pasukonis, Jimmy Ba, and Timothy Lillicrap.
\newblock Mastering diverse domains through world models.
\newblock {\em arXiv preprint arXiv:2301.04104}, 2023.

\bibitem{oudeyer2007typology}
Pierre-Yves Oudeyer and Frederic Kaplan.
\newblock What is intrinsic motivation? a typology of computational approaches.
\newblock {\em Frontiers in neurorobotics}, 1:6, 2007.

\bibitem{haber2018learning}
Nick Haber, Damian Mrowca, Stephanie Wang, Li~F Fei-Fei, and Daniel~L Yamins.
\newblock Learning to play with intrinsically-motivated, self-aware agents.
\newblock {\em Advances in neural information processing systems}, 31, 2018.

\bibitem{sekar2020planning}
Ramanan Sekar, Oleh Rybkin, Kostas Daniilidis, Pieter Abbeel, Danijar Hafner, and Deepak Pathak.
\newblock Planning to explore via self-supervised world models.
\newblock In {\em ICML}, 2020.

\bibitem{kauvar2023curious}
Isaac Kauvar, Chris Doyle, Linqi Zhou, and Nick Haber.
\newblock Curious replay for model-based adaptation.
\newblock {\em International Conference on Machine Learning}, 2023.

\bibitem{wang2016learning}
Jane~X Wang, Zeb Kurth-Nelson, Dhruva Tirumala, Hubert Soyer, Joel~Z Leibo, Remi Munos, Charles Blundell, Dharshan Kumaran, and Matt Botvinick.
\newblock Learning to reinforcement learn.
\newblock {\em arXiv preprint arXiv:1611.05763}, 2016.

\bibitem{wang2018prefrontal}
Jane~X Wang, Zeb Kurth-Nelson, Dharshan Kumaran, Dhruva Tirumala, Hubert Soyer, Joel~Z Leibo, Demis Hassabis, and Matthew Botvinick.
\newblock Prefrontal cortex as a meta-reinforcement learning system.
\newblock {\em Nature neuroscience}, 21(6):860--868, 2018.

\bibitem{maturana1987tree}
Humberto~R Maturana and Francisco~J Varela.
\newblock {\em The tree of knowledge: The biological roots of human understanding.}
\newblock New Science Library/Shambhala Publications, 1987.

\bibitem{maturana1970biology}
Humberto Maturana.
\newblock {\em Biology of cognition}.
\newblock Biological Computer Laboratory, Department of Electrical Engineering~{\ldots}, 1970.

\bibitem{varela1979principles}
Francisco~J Varela.
\newblock {\em Principles of biological autonomy}.
\newblock General Systems Research. North Holland, 1979.

\bibitem{froese2010life}
Tom Froese and Stewart John.
\newblock Life after ashby: ultrastability and the autopoietic foundations of biological autonomy.
\newblock {\em Cybernetics and Human Knowing}, 17(4):7--50, 2010.

\bibitem{rosen1971some}
Robert Rosen.
\newblock Some realizations of (m, r)-systems and their interpretation.
\newblock {\em The bulletin of mathematical biophysics}, 33:303--319, 1971.

\bibitem{hirota2023reformalizing}
Ryuzo Hirota, Hayato Saigo, and Shigeru Taguchi.
\newblock Reformalizing the notion of autonomy as closure through category theory as an arrow-first mathematics.
\newblock In {\em ALIFE 2023: Ghost in the Machine: Proceedings of the 2023 Artificial Life Conference}. MIT Press, 2023.

\bibitem{hamon2024discovering}
Gautier Hamon, Mayalen Etcheverry, Bert Wang-Chak Chan, Cl{\'e}ment Moulin-Frier, and Pierre-Yves Oudeyer.
\newblock Discovering sensorimotor agency in cellular automata using diversity search.
\newblock {\em arXiv preprint arXiv:2402.10236}, 2024.

\bibitem{ganti2003chemoton}
Tibor G{\'a}nti.
\newblock {\em Chemoton theory: theory of living systems}.
\newblock Springer Science \& Business Media, 2003.

\bibitem{luisi2003autopoiesis}
Pier~Luigi Luisi.
\newblock Autopoiesis: a review and a reappraisal.
\newblock {\em Naturwissenschaften}, 90:49--59, 2003.

\bibitem{friston2013life}
Karl Friston.
\newblock Life as we know it.
\newblock {\em Journal of the Royal Society Interface}, 10(86):20130475, 2013.

\bibitem{bruineberg2022emperor}
Jelle Bruineberg, Krzysztof Do{\l}{\k{e}}ga, Joe Dewhurst, and Manuel Baltieri.
\newblock The emperor's new markov blankets.
\newblock {\em Behavioral and Brain Sciences}, 45:e183, 2022.

\bibitem{araki2013long}
Takaya Araki, Tomoaki Nakamura, and Takayuki Nagai.
\newblock Long-term learning of concept and word by robots: Interactive learning framework and preliminary results.
\newblock In {\em 2013 IEEE/RSJ International Conference on Intelligent Robots and Systems}, pages 2280--2287. IEEE, 2013.

\bibitem{taniguchi2022whole}
Tadahiro Taniguchi, Hiroshi Yamakawa, Takayuki Nagai, Kenji Doya, Masamichi Sakagami, Masahiro Suzuki, Tomoaki Nakamura, and Akira Taniguchi.
\newblock A whole brain probabilistic generative model: Toward realizing cognitive architectures for developmental robots.
\newblock {\em Neural Networks}, 150:293--312, 2022.

\end{thebibliography}

\newpage
\appendix

\section*{Acknowledgment}
We thank Ignacio Cea for valuable comments on an earlier version of the manuscript. This research is supported by Japan Society for the Promotion of Science KAKENHI grant 24K23892 for NY and 24K20859 for KH.

\section{Mini Review for Autopiesis}
\label{sec:review}

This review provides a comprehensive overview of the development of the concept of autopoiesis since its inception 50 years ago, and its computational models. The concept of autopoiesis, first introduced by Chilean biologists Humberto Maturana and Francisco Varela in the early 1970s \cite{maturana1987tree,maturana1970biology, maturana1980autopoiesis, varela1979principles}, has had a profound impact on our understanding of life, cognition, and complex systems. Autopoiesis refers to the process by which self-maintaining and self-producing systems sustain themselves and maintain their identity through continuous interactions with their environment. The concept of autopoiesis later became connected with ideas such as enaction related to the sensorimotor loop \cite{di2014enactive} and agency \cite{barandiaran2009defining}, and biological autonomy \cite{froese2010life, moreno2015biological}.

We then delve into the diverse research streams that have emerged in the field as computational model, examining the Category theory approach, Enactive approach, Synthetic Biology approach, and Bayesian approach each with its unique contributions to the understanding of autopoiesis. 
First, there is the approach using category theory. This begins with Rosen's (M,R) system \cite{rosen1971some}, and more recently, discussions on closure have been conducted by Moreno and Mossio \cite{moreno2015biological}, and Hirota, Saigo and Taguchi\cite{hirota2023reformalizing}. 
Next is the Enactive Approach. While Di Paolo and Frose have conceptually organized the interactions between agents and their environment \cite{di2014enactive, froese2010life}, computational models like the Sensorimotor Lenia \cite{hamon2024discovering}, which employs cellular automata, have been proposed. 
The third is the Synthetic Biology approach, which began with Ganti's chemoton\cite{ganti2003chemoton} and has been further modeled by Luisi \cite{luisi2003autopoiesis}. Lastly, there is the formulation of autopoietic systems using the free energy principle and Markov blankets \cite{friston2013life, bruineberg2022emperor}.Autopoiesis has developed both conceptually and computationally over the past 50 years, serving as an important guideline for constructing artificial agency.

\section{Graphical abstract}
\label{sec:gabst}

\begin{figure}[H]
  \centering
  \includegraphics[width=0.8\linewidth,bb=0 0 695 307]{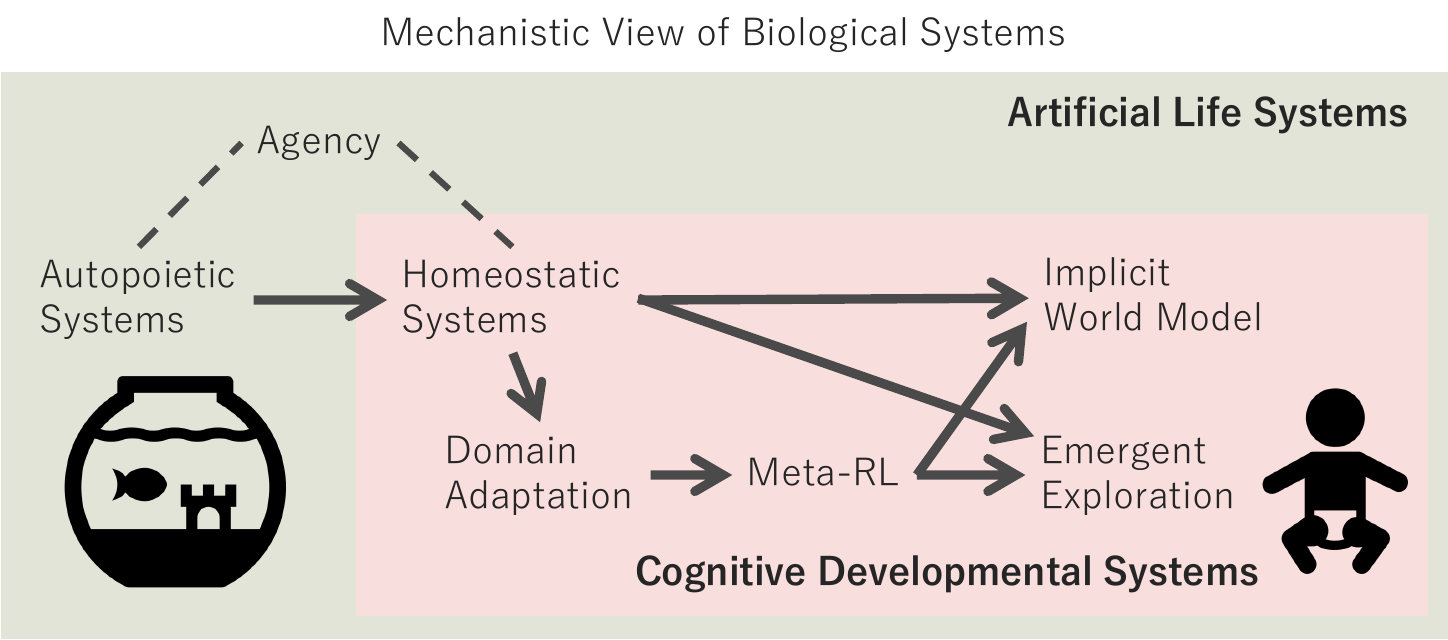}
  \caption{Relation diagram of our proposal on the emergent abilities of autonomous cognitive developmental systems, from mechanistic (= undirected, unsupervised) perspective of autonomous biological systems. }
  \label{fig:autopoiesis}
\vspace{-4mm}
\end{figure}

\section{Architecture for Mortal Agent}
\label{sec:meta_hrl}

\begin{figure}[H]
  \centering
  \includegraphics[width=0.8\linewidth,bb=0 0 815 444]{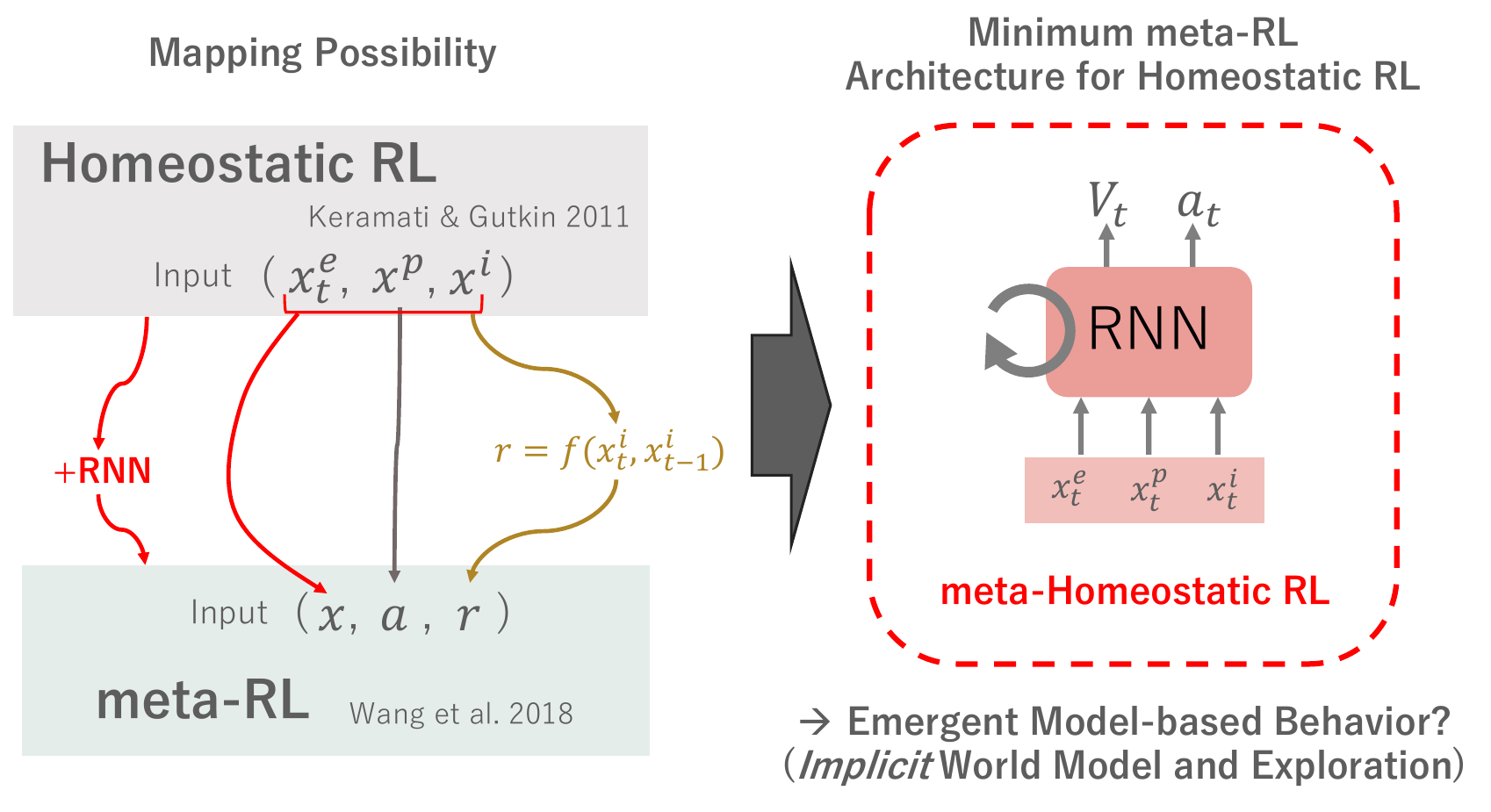}
  \caption{Implication of homeostatic extrinsic reward system combined with recurrent connection for the emergence of implicit world models and exploration.}
  \vspace{-5mm}
  \label{fig:metahrl}
\end{figure}

By combining recent meta-RL \cite{wang2016learning,wang2018prefrontal} and deep homeostatic RL, we propose the possibility of explaining IM as an emergent property of systems adapting to a domain, together with a world model. To do this, we first focus on the possibility of mapping meta-RL and homeostasis RL (Figure \ref{fig:metahrl}). Specifically, the external observations $x_t$, latest action selection $a_{t-1}$, and latest reward $r_{t-1}$ required for domain adaptation in meta-RL. These multi-modal observation are thought to correspond to exteroception $x^e$, proprioception $x^p$, and interoception $x^i$, in homeostatic RL \cite{yoshida2024emergence} respectively. The multi-modal observation is common situation in studies of cognitive developmental robotics \cite{araki2013long,taniguchi2022whole}. Therefore, the inclusion of a recurrent neural networks (RNNs), which is essential for meta-RL, in the model architecture  of homeostasis RL agent may minimally lead to the potential for a meta-learning ability. 

\end{CJK}
\end{document}